%% file: ms.tex.tex
\RequirePackage{fix-cm} 

\documentclass[article,twocolumn,apacite,floatsintext]{apa6}

\title{Optimality and limitations of audio-visual integration for cognitive systems}
\author{W. Paul Boyce$^{1,+,*}$, Tony Lindsay$^{1}$, Arkady Zgonnikov$^{2,3}$, Ignacio Rano$^{1,\alpha}$, \& KongFatt Wong-Lin$^{1}$}

\affiliation{$^{1}$Intelligent Systems Research Centre, Ulster University, Magee campus, Derry~Londonderry, Northern Ireland, UK\\$^{2}$AiTech, Delft University of Technology, Delft, The Netherlands\\$^{3}$Department of Cognitive Robotics, Faculty of Mechanical, Maritime, and Materials Engineering, Delft University of Technology, Delft, The Netherlands\\$^{+}$WPB is now at the School of Psychology, University of New South Wales, Sydney, NSW, Australia.\\$^{\alpha}$IR is now at SDU Biorobotics, Mærsk Mc Kinney Møller Institut, University of Southern Denmark, Denmark.\\$^{*}$Correspondence: W. Paul Boyce\\William.Boyce@unsw.edu.au}

\renewcommand{\abstractname}{\large Abstract}

\shorttitle{Optimality and limitations of audio-visual integration for cognitive systems}
\leftheader{Boyce, Lindsay, Zgonnikov, Rano, \& Wong-Lin}

\usepackage{etoolbox}

\patchcmd{\subsection}{\bfseries}{\relax}{}{}

\usepackage{lipsum}

\usepackage{apacite}
\usepackage{fixltx2e} 
\usepackage{color,soul} 

\usepackage{caption}
\usepackage{txfontsb}
\usepackage{verbatim}
\usepackage{epstopdf}

\usepackage{placeins} 

\usepackage{url}
\usepackage[english]{babel} 

\usepackage{setspace}
\usepackage{lettrine} 
\usepackage{tabulary}

\usepackage{lscape}	

\begin{document}

\maketitle

\begin{abstract}

\textbf{\textit{\large \abstractname}--Multimodal integration is an important process in perceptual decision-making. In humans, this process has often been shown to be statistically optimal, or near optimal: sensory information is combined in a fashion that minimises the average error in perceptual representation of stimuli. However, sometimes there are costs that come with the optimization, manifesting as illusory percepts. We review audio-visual facilitations and illusions that are products of multisensory integration, and the computational models that account for these phenomena. In particular, the same optimal computational model can lead to illusory percepts, and we suggest that more studies should be needed to detect and mitigate these illusions, as artefacts in artificial cognitive systems. We provide cautionary considerations when designing artificial cognitive systems with the view of avoiding such artefacts. Finally, we suggest avenues of research towards solutions to potential pitfalls in system design. We conclude that detailed understanding of multisensory integration and the mechanisms behind audio-visual illusions can benefit the design of artificial cognitive systems.}

\end{abstract}


\section{Introduction}
Perception is a coherent conscious representation of stimuli that is arrived at, via processing signals sent from various modalities, by a perceiver: either human or non-human animals \cite{Goldstein2010}. Humans have evolved multiple sensory modalities, which include not only the classical five (sight, hearing, tactile, taste, olfactory) but also more recently defined ones (for example, time, pain, balance, and temperature \cite{Rao2001evolution, Fulbright2001functional,Fitzpatrick1994proprioceptive, Green2004temperature}). While each modality is capable of resulting in a modality-specific percept, it is often the case that stimulus information gathered by two or more modalities is combined in an attempt to create the most robust representation possible of a given environment in perception \cite{Macaluso2000modulation, Ramos2007visual}.

Understanding and mapping just how the human brain combines different types of stimulus information from drastically different modalities is challenging. Behavioural studies have suggested optimal or near-optimal integration of multi-modal information \cite{Alais2004ventriloquist, Shams2010}. In the case of \citeA{Alais2004ventriloquist} they examined the classic spatial ventriloquist effect through the lens of near optimal binding. The effect in question describes the apparent `capture' of an auditory stimulus in perceptual space that is then mapped to the perceptual location of a congruent visual stimulus, the famous example being the ventriloquist's voice appearing to emanate from the synchronously animated mouth of the dummy on their knee. \citeA{Alais2004ventriloquist} demonstrated that this process of `binding' the perceived spatial location of an auditory stimulus to the perceived location of a visual stimulus is an example of near optimal audio-visual integration. They achieved this by demonstrating that variations of the effect could be reversed (i.e. a visual stimulus being `captured' and shifted to the same perceptual space as an auditory stimulus) when the auditory signal was less noisy relative to the visual stimulus (when extreme blurring noise was added to the visual stimulus). Additionally, when visual stimuli was blurred, but not extremely so, neither stimulus source captured the other and a mean spatial position was perceived. This in turns hints at a weighting process in audio-visual integration modulated by the level of noise in a given source signal. These findings are consistent with the notion of inverse effectiveness: when a characteristic of a given stimulus has low resolution there tends to be in an increase in `strength' of multisensory integration \cite{de2017effects, stevenson2009audiovisual}. See \citeA{holmes2009principle} for potential issues when measuring multisensory integration `strength' from the perspective of inverse effectiveness.

However, the very existence of audio-visual illusions in these processes highlights that there can be a cost associated with this optimal approach \cite{Shams2005optimal}; the perceptual illusions here are being considered as unwanted artefacts (costs) that manifest due to optimal integration of signals from multiple modalities. One such well established audio-visual illusion that combines information from both modalities and arrives at an auditory percept altogether unique is the McGurk-MacDonald effect \cite{McGurkMacDonald1976}. When participants watch footage of someone moving their lips, while simultaneously listening to an auditory stimulus (a single syllable repeated in time with the moving lips) that is incongruent to the moving lips, they have a tendency to `hear' a sound that matches neither the mouthed syllable or the auditory stimulus. While not gazing at the moving lips, participants accurately report the auditory stimulus.

The McGurk effect demonstrates that the integration of audio-visual information is an effective process in most natural settings (even when modalities provide competing information), but may occasionally result in an imperfect representation of events. This auditory illusion suggests a ``best guess'' can sometimes be arrived at when modalities provide contradictory information, where different weightings are given to competing modalities. \citeA{Crevecoeur2016} highlighted that the nervous system also considers temporal feedback delays when performing optimal multi-sensory integration (for example, visual input followed by muscle response is slower than proprioceptive input followed by a muscle response with a difference of \textasciitilde50ms). The faster of the two sensory cues is given a dominant weighting in integration. This shows that temporal characteristics affect optimal integration of information from different modalities, and should be a factor in any models of multi-modal integration.

If artefacts such as illusions can occur in an optimal multi-modal system, these artefacts become a concern when designing artificial cognitive systems. The optimal approach of integrating information from multiple sources may lead to inaccurate representation of an environment (an artefact), which in turn could result in a potentially hazardous outcome. For example, if a autonomous vehicle was trained in a specific environment and then relocated to a novel environment, an artefact manifested via optimal integration of stimuli could compromise the safe navigation through the novel environment and any action decisions taken therein (this scenario is a combination of the ``Safe Exploration'' and ``Robustness to Distributional Shift'' accident risks as outlined by \citeA{Amodei2016concrete}).

The remainder of the paper reviews the processes in audio-visual perception that offers explanations for audio-visual illusions and effects, focusing mainly on how audition affects visual perception, and what this tells us about the audio-visual integration system. We continue by building a case for audio-visual integration as a process of evidence accumulation/discounting, where differing weights are given to different modalities depending on the stimuli information (spatial, temporal, featural etc.) being processed, which follows a hierarchical process (from within-modality discrimination to multi-modal integration). We highlight how some processes are optimal and others suboptimal, and how each have their own drawbacks. Following that, we review cognitive models of multi-modal integration which provide computational accounts for illusions. We then outline the potential implications of the mechanisms behind multisensory illusions for artificial intelligent systems, concluding with our views on future research directions. Additionally, rather than assuming that all attributions of prior entry (discussed below) are accurate, this paper expands on the definition of prior entry to encompass both response bias and undefined non-attentional processes. Doing so circumvents the granular debates surrounding prior entry in favour of better discussing the broader processes on the way to audio-visual integration, of which prior entry is but one. We also consider impletion (discussed below) as a process distinct from prior entry, but one that complements and/or competes with prior entry.

\section{Audio-visual Integration}

\subsection{Visual and Multi-modal Prior Entry}
Prior entry, a term coined by E.B. Titchener in 1908, describes a process whereby a visual stimulus that draws an observer's attention is processed in the visual perceptual system before any unattended stimuli in the perceptual field. This in turn results in the attended stimulus being processed ``faster'' relative to subsequently attended stimuli \cite{SpenceKelinShore2001}. This suggests that when attention is drawn (usually via a cue) to a specific region of space, a stimulus that is presented to that region is processed at a greater speed than a stimulus presented to unattended space. 

Prior entry as a phenomenon is important in multi-modal integration due to the fact that the temporal perception of one modality can be significantly altered by stimuli in another modality (as well as within a modality) \cite{Shimojo1997}. The mechanisms underlying prior entry have been the subject of controversy \cite{Cairney1975, Schneider2003, Downing1997}, but strong evidence has been provided for its existence via orthogonally designed crossmodal experiments \cite{SpenceKelinShore2001, Zampini2005}. In the case of the orthogonally designed experiments, related information between the attended modality and the subsequent temporal order judgement task was removed, thus ensuring no modality-specific bias.

\begin{figure}[h!]
	\begin{center}
		\includegraphics[trim=190 210 250 150,clip,scale=.7, keepaspectratio]{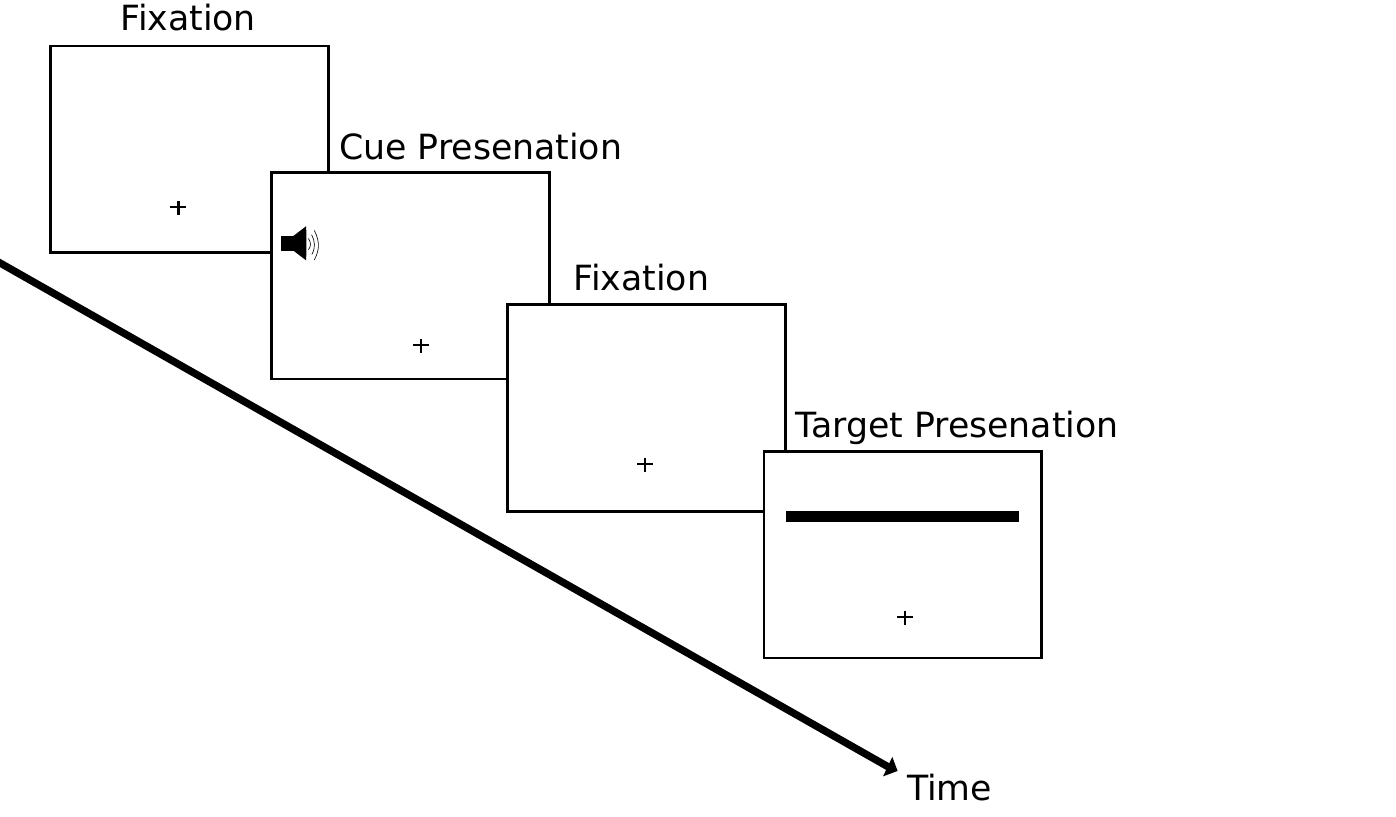}
	\end{center}
	\captionof{figure}{\label{fig:LineIllusion}The paradigm of the line-motion illusion similar to that used by \protect\citeA{Hikosaka1993a} and \protect\citeA{Shimojo1997}. Participants were presented with a fixation cross followed by a cue (auditory in this example) to one side of fixation followed by another fixation display. Finally the target stimulus was presented. In this example the left side of the target stimulus was cued via an auditory tone. The resulting perception would be that of the line being drawn from the same side as the cue as opposed to being presented in its entirety at the same time. }
\end{figure}

A classic visual illusion that supports the tenets of prior entry, and demonstrates just how much temporal perception can be affected by it, is the line motion illusion, first demonstrated by \citeA{Hikosaka1993a} using visual cues. A cue to one side of fixation prior to the presentation of a whole line to the display can result in the illusion of the line being ``drawn'' from the cued side (Figure~\ref{fig:LineIllusion}). \citeA{Hikosaka1993} investigated this effect further and demonstrated illusory temporal order in a similar fashion: namely, cueing one side of fixation in a temporal order judgement task prior to the simultaneous onset of both visual targets. Both these effects were replicated using auditory cues by \citeA{Shimojo1997}.

\citeA{Shimojo1997} demonstrated that the integration of auditory and visual stimuli can cause temporal order perception in one modality to be significantly altered by information in another via audio-visual prior entry. However, \citeA{Downing1997} suggested that the original line motion illusion was an example of what they termed ``impletion'': in an ambiguous display multiple stimuli are combined to reflect a single smooth event in perception. For example, when an illusion of apparent motion is created using statically flashed stimuli in different locations (e.g. left space followed by right space), the stimuli can appear to smoothly change from the first stimulus shape (circle) to the second stimulus shape (square) \cite{Kolers1976shape, Downing1997}. It is suggested that a discriminatory process gathers all available stimuli information, combines them, and creates a coherent percept; or the most likely real world outcome where it fills in the gaps on the way to perception. \citeA{Downing1997} demonstrated that the line motion illusion could in fact be a perception of the visual cue itself streaking across the field of display akin to frames in an animation. Admittedly, when one takes into account the findings of \citeA{Shimojo1997} using auditory cues, it may be tempting to dismiss the account of impletion, but illusory visual motion can be induced via auditory stimuli \cite{Hidaka2009}, which demonstrates that auditory stimuli can also induce a perception of motion in visual modality independent of prior-entry. Despite these alternative explanations for phenomena such as the line motion illusion, neuroscience has provided strong evidence for the existence of prior entry: speeded processing when attention was directed to the visual modality rather than the tactile \cite{Vibell2007}, speeded processing associated with attending to an auditory stimulus \cite{Folyi2012}, and speeded processing during a visual task when an auditory tone was presented prior to the onset of the visual stimuli \cite{Seibold2014}.

Evidence thus suggests that prior entry, and indeed audio-visual prior entry, is a robust phenomenon. Whether all effects attributed to prior entry are done so correctly is another matter, but ultimately may be somewhat irrelevant (see \citeA{Fuller2009} where evidence for both prior entry and impletion in the line motion illusion is presented, and suggests prior entry is not requisite). For instance, even if response bias or some non-attentional processes are mistakenly attributed to prior entry, these effects are still predictable, and replicable, and in fact these processes may enhance or exaggerate genuine prior entry effects.

The prior entry and impletion effects discussed above show that shifts in attention, or the combination of separate stimuli into the perception of a single stimulus event, can result in illusory temporal visual perception. It seems likely that evidence gathered from both the audio and visual modalities are combined optimally with some sources of information being given greater weighting in this process. When and how to assign weightings in an artificial system is an important consideration in design in order to avoid artefacts such as those described above. While prior entry and/or impletion can result in an inaccurate representation of temporal events, there exist audio-visual effects that are facilitatory in nature and thus desirable, which we discuss next.
\vspace{1em}

\subsection{Temporal ventriloquism}

\begin{figure}[h!]
	\begin{center}
		\includegraphics[scale=.6,keepaspectratio]{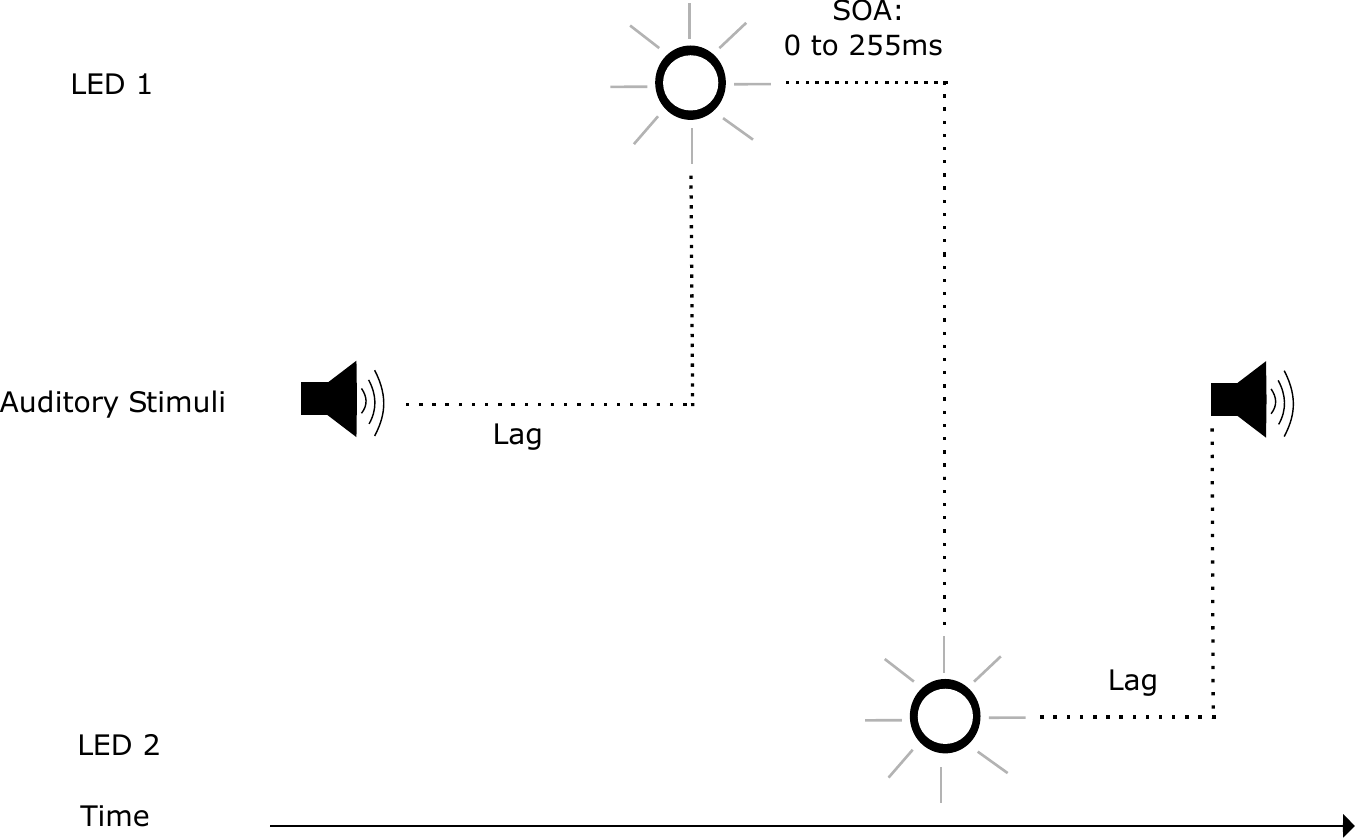}
	\end{center}
	\captionof{figure}{\label{fig:TOJ_Enhancement}Trial events in the temporal ventriloquism paradigm that resulted in enhanced temporal order judgement performance \protect\cite{Morein-Zamir2003}. The stimulus onset asynchrony (SOA) between illumination of LEDs varied. The first auditory tone was presented before the first LED illumination. The second auditory tone was presented after the second LED illumination. The SOA between tones also varied.}	
\end{figure}

\begin{figure}[h!]
	\begin{center}
		\includegraphics[scale=.6, keepaspectratio]{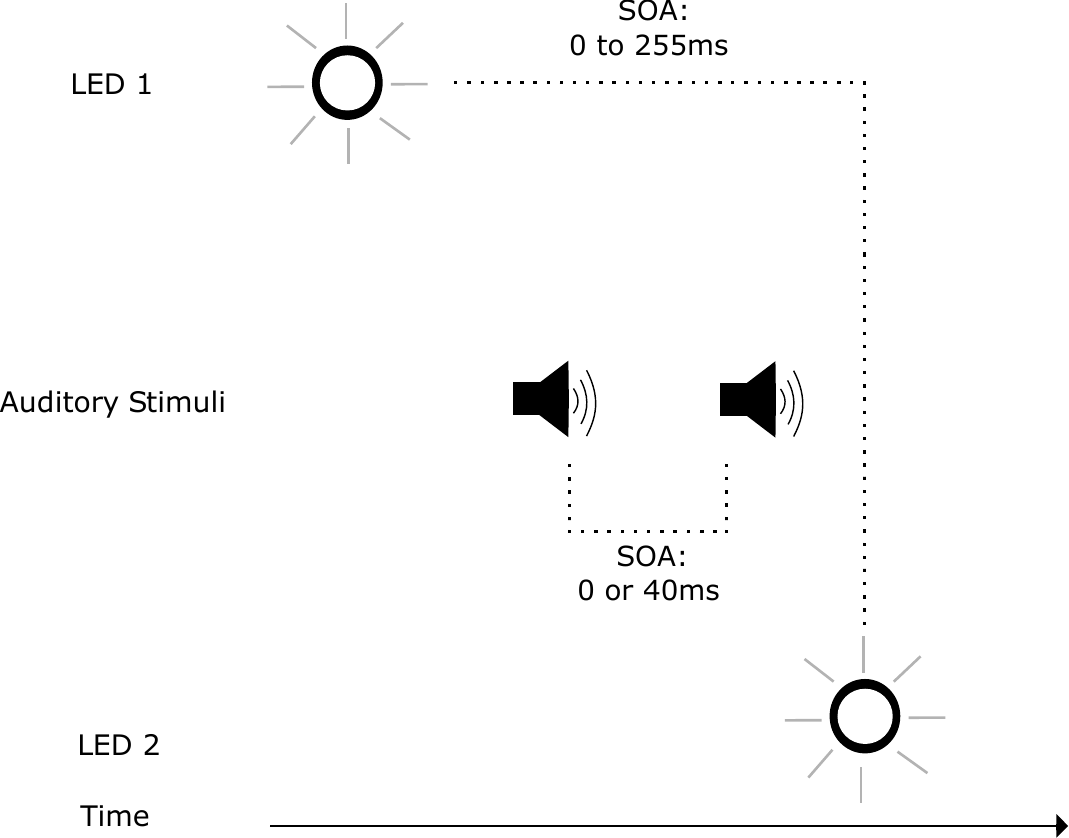}
	\end{center}
	\captionof{figure}{\label{fig:TOJ_Detrimment}Trial events in the temporal ventriloquism paradigm that resulted in a detriment in temporal order judgement performance \protect\cite{Morein-Zamir2003}. The stimulus onset asynchrony (SOA) between illumination of LEDs varied. The first auditory tone was presented after the first LED illumination. The second auditory tone was presented before the second LED illumination. The SOA between tones also varied.}
\end{figure}

Illusory visual temporal order, as shown above, can be induced by auditory stimuli. However, auditory stimuli, when integrated with visual stimuli, can also facilitate visual temporal perception:	\citeA{Scheier1999} discovered an audio-visual effect where spatially non-informative auditory stimuli affected the temporal perception of a visual temporal order judgement task. This effect became known as temporal ventriloquism, analogous to spatial ventriloquism where visual stimuli shifts the perception of auditory localisation \cite{Radeau1987, Willey1937, Thomas1941}. Temporal ventriloquism was further investigated by \citeA{Morein-Zamir2003}: when accompanied by auditory tones, performance in a visual temporal order judgement task was enhanced (Figure~\ref{fig:TOJ_Enhancement}). This enhancement was abolished when the tones coincided with the visual stimuli in time. When the two tones were presented between the visual stimuli in time (Figure~\ref{fig:TOJ_Detrimment}), a detriment in performance was observed. In both conditions the tones appeared to ``pull'' the perception of the visual stimuli in time towards the auditory stimuli temporal onsets: further apart in the enhanced performance and closer together when a detriment in performance was observed (see however \citeA{Hairston2006auditory} for an argument against the notion of introducing a temporal gap between the stimuli in visual perception). The main driver of this effect was believed to be the temporal relationship between the auditory and visual stimuli, where the higher temporal resolution of the auditory stimuli carried more weight in integration. This is a potent example of how assigning weightings in multi-modal integration can have both positive and negative outcomes in terms of system performance.

\citeA{Morein-Zamir2003} hypothesized that the quantity of auditory stimuli must match the quantity of visual stimuli in order for the temporal ventriloquism effects to occur. For example, when a single tone was presented between the presentation of the visual stimuli in time, there was no reported change in performance. \citeA{Morein-Zamir2003} refer to the unity assumption: the more physically similar stimuli are to each other across modalities, the greater the likelihood they will be perceived as having originated from the same source \cite{Welch1999}, we discuss this in more detail later.

However, other research questions the assumption that a matching number of stimuli in both the auditory and visual modality are required to induce temporal ventriloquism. \citeA{Getzmann2007} studied an apparent motion paradigm, where participants perceive two sequentially presented visual stimuli behaving as one stimulus moving from one position to another. They found that when a single click was presented between the two visual stimuli, it increased the perception of apparent motion, essentially ``pulling'' the visual stimuli closer together in time in perception. This casts doubt on the idea that the quantity of stimuli must be equal across modalities in order for, in this instance, an auditory stimulus to affect the perception of visual events. 

Indeed, \citeA{Boyce2019} demonstrated that a detriment in response bias corresponding to actual visual presentation order can be achieved with the presentation of a single tone in neutral space (different space to the visual stimuli) in a visual ternary response task (temporal order judgement combined with simultaneity judgements where the participant reports if stimuli were presented simultaneously). Importantly, this can be achieved consistently when presenting the single tone \textit{prior} to the onset of the first visual stimulus (similar to the trial in Figure \ref{fig:TOJ_Enhancement} but without the second auditory stimulus). Often participants were as likely to make a simultaneity judgement report as they were to make a temporal order judgement report corresponding to actual sequential visual stimuli presentation order. This poses a problem for the temporal ventriloquism narrative: a single tone before the sequential presentation of visual stimuli in a ternary task would be expected to ``pull'' the perception of the first visual stimulus towards it in time, resulting in increased reports corresponding to the sequential order of visual stimuli. Alternatively, it might ``pull'' the perception of both visual stimuli in time with no observable effect on report bias should the matching quantity rule be abandoned. The repeated demonstrations of a decrease in report bias corresponding to the sequential order of visual stimuli suggest that the processes underlying temporal ventriloquism may be more flexible than previously suggested, and may have impletion-esque elements. Specifically, characteristics of stimuli such as their spatial and temporal relationship, and the featural similarity of stimuli within a single modality, may be weighted to arrive at the most likely real world outcome in perception regardless of whether the number of auditory stimuli equal the number of visual stimuli or not. Indeed, the number of auditory stimuli relative to visual in this example appears to modulate the type of temporal ventriloquist effect that might be expected to be observed.

\begin{figure*}[h!]
	\begin{center}
		\includegraphics[scale=.6, keepaspectratio]{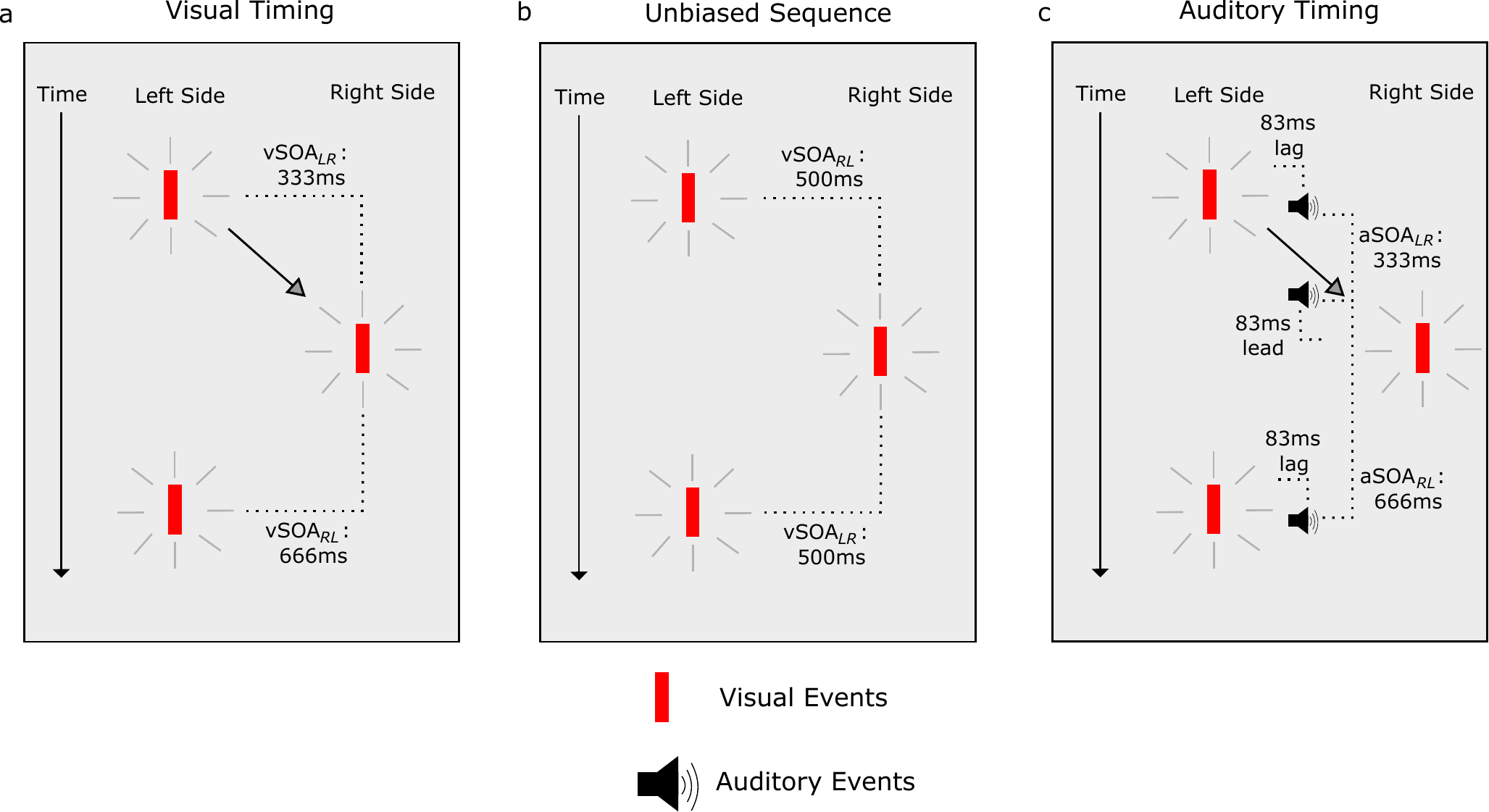}
	\end{center}
	\captionof{figure}{\label{fig:Apparent_Motion} Illusory apparent visual motion paradigm \protect\cite{Freeman2008}; \textbf{(a)} the visual SOAs (vSOA) between stimuli. `L' denotes the left stimulus, and `R' the right stimulus. When the vSOA was $333$ms, apparent motion in the direction of the second stimulus was perceived. When there was a vSOA of $666$ms, no apparent motion was perceived; \textbf{(b)}: when vSOAs were $500$ms, there was no bias in apparent motion. \textbf{(c)}: when vSOAs were $500$ms but auditory stimuli were presented with an SOA (aSOA) of $333$ms, participants perceived apparent motion in the same direction as would be expected with a vSOA of $333$ms. When the aSOA was $666$ms, illusory apparent visual motion was abolished. }
	\vspace{-1.5em}
\end{figure*}

Clearly not all conditions support the idea that the temporal relationship of an auditory stimulus to a visual stimulus drives temporal ventriloquism and similar effects. However, while there are no easy explanations for the audio-visual integration in temporal ventriloquism, efforts have been made to show that auditory stimuli do indeed ``pull'' visual stimuli in temporal perception. \citeA{Freeman2008} created an innovative paradigm that tested the idea that temporal ventriloquism is driven by auditory capture (in a similar fashion to that of \citeA{Getzmann2007}), that is to say there is a ``pulling'' of visual stimuli towards auditory stimuli in temporal perception. They began by determining the relative timings of visual stimuli that resulted in illusory apparent visual motion (Figure \ref{fig:Apparent_Motion}). Once visual stimulus onset asynchronies (SOAs) were established for the effect, \citeA{Freeman2008} adjusted the timings to remove bias in the illusion (Figure \ref{fig:Apparent_Motion}b). Following that, they introduced auditory stimuli (Figure \ref{fig:Apparent_Motion}c) with the same SOAs used to induce the effect in the visual-only condition (Figure \ref{fig:Apparent_Motion}a). In the presence of the auditory stimuli, both visual stimuli were ``pulled'' towards each other in time perception, and participants perceived a bias in the illusion. 

This demonstrated that auditory stimuli had the ability to ``pull'' the respective visual stimuli in perceptual time towards the respective auditory onsets. In doing so, the visual stimuli now appeared in perception to have the same SOA as the auditory stimuli. This introduced a perceptual bias consistent with that observed for the visual stimuli SOA (in the absence of  auditory stimuli) necessary to induce the same bias in illusory apparent visual motion. This meant predictable manipulation of the effect, and more specifically, demonstrated auditory capture of visual events in time.

\citeA{Freeman2008} suggest that the timings of the flanker stimuli (the stimuli used to induce temporal ventriloquism effects) in relation to the visual are the main drivers of temporal ventriloquism. \citeA{Roseboom2013a} demonstrated that, in fact, the featural similarity of the flanker stimuli used to induce the effects described by \citeA{Freeman2008} have arguably as important a role to play at these time scales. Specifically, \citeA{Roseboom2013a} replicated the findings of \citeA{Freeman2008} using auditory flankers. When flankers were featurally distinct (e.g. a sine wave and a white-noise burst) or flanker types were mixed via audio-tactile flankers, a mitigated effect was observed. It was significantly weaker compared to featurally identical audio-only or tactile-only flankers. This suggests that temporal capture in-and-of-itself is not sufficient when describing the underlying mechanisms that account for this effect, or temporal ventriloquism in general at the reported time scales. \citeA{Roseboom2013a} also demonstrated that the reported illusory apparent visual motion could be induced when the flanker stimuli was presented synchronously with the target visual stimuli. This suggests that temporal ansynchrony is not a requisite to induce this illusion in a directionally ambiguous display. \citeA{Keetels2007} further highlighted the importance of featural characteristics when inducing the temporal ventriloquism effect. However, at shorter time scales, featural similarity appears not to play as large a role where timing is reasserted as the main driver \cite{Klimova2017grouping, Kafaligonul2010auditory, Kafaligonul2012static}. 

The above research is consistent with the unity assumption, where an observer makes an assumption about two sensory signals, such as auditory and visual (or indeed, signals from the same modality), originating from a single source or event \cite{vatakis2007crossmodal, vatakis2008evaluating, chen2017assessing}. \citeA{vatakis2007crossmodal} demonstrated that when auditory and visual stimuli were mismatched (for example, speech presentation where the voice did not match the gender of the speaker) participants found it easier to judge which stimulus was presented first; auditory or visual. The task difficulty increased when the stimuli were matched suggesting an increased likelihood of perceiving the auditory and visual stimuli occurring at the same time. This finding provides support for the unity assumption in audio-visual temporal integration of speech via the process of temporal ventriloquism. See \citeA{vatakis2008evaluating} for limitations of the unity assumption's influence over audio-visual temporal integration of complex non-speech stimuli. See also \citeA{chen2017assessing} for a thorough review of the unity assumption and the myriad debates surrounding it, and how it relates to Bayesian causal inference.

The findings by \citeA{Roseboom2013a} and \citeA{Keetels2007} show that there is often a much more complex process of integration than simply auditory stimuli (or other stimuli of high temporal resolution) capturing visual stimuli in perception. There would appear to be a process of evidence accumulation and evidence discounting: when two auditory events are featurally similar, and their temporal relationship with visual stimuli is close, the auditory and visual stimuli are integrated. However, when two auditory stimuli meet the temporal criteria for integration with visual stimuli, but these auditory stimuli are featurally distinct and therefore deemed to be from unique sources, they are not wholly integrated with the visual stimuli. In the second example, some of the accumulated temporal evidence is discounted due to evidence of unique sources being present.

Taken together, this evidence builds a more complicated picture of temporal ventriloquism in general, and the modulated illusory apparent visual motion direction effect~\cite{Freeman2008} in particular. Indeed, the unity assumption potentially plays a role here when one considers the effect featural similarity has on the apparent grouping of flankers.
\vspace{1em}

\subsection{Additional audio-visual effects}
\begin{figure}[h!]
	\begin{center}
		\includegraphics[scale=.8, keepaspectratio]{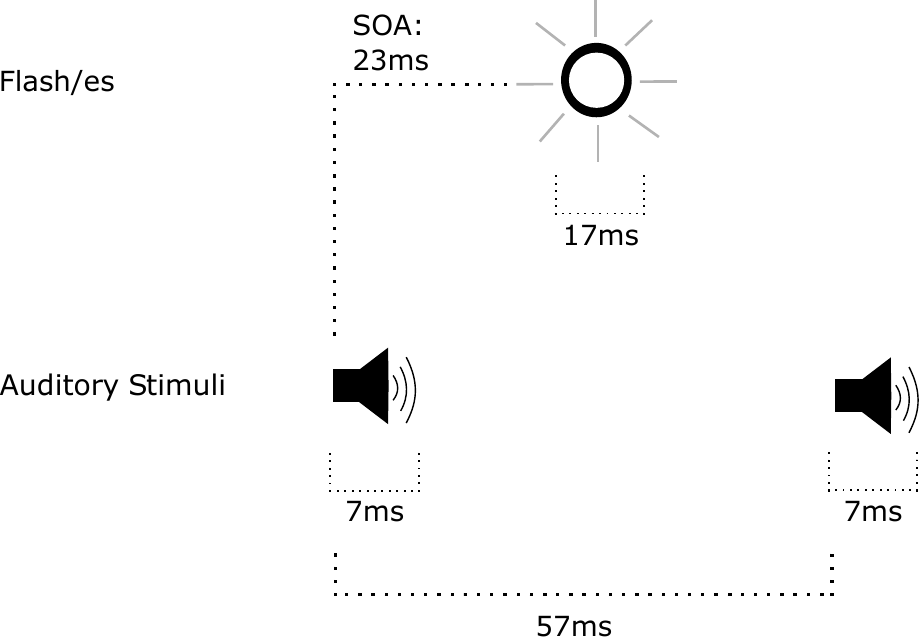}
	\end{center}
	\captionof{figure}{\label{fig:DFI}Trial events in the multiple flash illusion paradigm \protect\cite{Shams2002}. A tone was presented before visual stimulus onset, and a tone was presented after visual stimulus onset. The SOA between tones was 57ms. This paradigm resulted in the perception of multiple flashes when in fact the visual stimulus was presented only once for 17ms. }
	
\end{figure}

Most of the research discussed thus far has focused on the effects of audio-visual integration on space and time perception in the visual modality. As highlighted by the McGurk effect, audio-visual integration can also have other surprising outcomes in perception. \citeA{Shams2002} demonstrated that when a single flash of a uniform disk was accompanied by two or more tones, participants tended to perceive multiple flashes of the disk (Figure \ref{fig:DFI}). When multiple physical flashes were presented and accompanied by a single tone, participants tended to perceive a single presentation of the disk \cite{Andersen2004}. These effects were labelled as \textit{fission} in the case of illusory flashes, and \textit{fusion} in the case of illusory single presentation of the disk. Fission and fusion differ from the likes of temporal ventriloquism and prior entry in that they increase or decrease the quantity of perceived stimuli. After training, or when there was a monetary incentive, qualitative differences were detectable between illusory and physical flashes \cite{vanErp2013, Rosenthal2009}. However, the illusion persisted despite the ability to differentiate. Similarities may be drawn between the effect reported by \citeA{Shams2002} and \citeA{Shipley1964} where, when the flutter rate of an auditory signal was increased, participants perceived an increased flicker frequency of a visual signal. However, there was a relatively small quantitative change in flicker frequency, whereas fission is a pronounced change in the visual percept (a single stimulus perceived as multiple stimuli).


Neuroimaging evidence provided further insights into fission/fusion effects. Specifically, in the presence of auditory stimuli the BOLD response in the retinotopic visual cortex increased whether fission was perceived or not \cite{Watkins2006}. The inverse was true when the fusion illusion was perceived. This suggests that the auditory and visual perceptual systems are intrinsically linked, and reflects the additive nature of the fission illusion and the suppressive nature of the fusion illusion that ``removes'' information from visual perception (auditory stimuli has also been shown to have suppressive effects on visual perception~\cite{Hidaka2015sound}).

\citeA{Shams2002} proposed the discontinuity hypothesis as an underlying explanation for the fission effect: discontinuous stimuli 

must be present in one modality in order to ``dominate'' another modality during integration. However, as alluded to above, \citeA{Andersen2004} demonstrated that this was not the case via the fusion illusion. 
Fission and fusion once again align with the ideas of impletion and the unity assumption. Consistent with the influence featural similarity of flankers had on illusory apparent visual motion, the fission effect was completely abolished when the tones used were distinct from each other: one a sine wave, the other a white noise burst; or both featurally distinct sine wave tones (a 300Hz sine wave and a 3500Hz) \cite{Roseboom2013b, BoycePhD2016}.

Another effect that seems to be governed by the featural similarity of auditory stimuli is the stream bounce illusion. In this illusion, two uniform circles move towards each other from opposite space, and when a tone presented at the point of overlap of the circles differed featurally to other presented tones, the circles are perceived to ``bounce'' off each other. When multiple tones were featurally identical, the circles often appeared to cross paths and continue on their original trajectory \cite{Sekuler1997sound, Watanabe2001}. Taken with the above and similar research \cite{Keetels2007, Cook2009}, this suggests that auditory streaming (where a sequence of auditory stimuli are assigned the same or differing origins) processes play an integral role in audio-visual illusions and integration in general.

\begin{figure}
	\begin{center}
		\includegraphics[scale=.7, keepaspectratio]{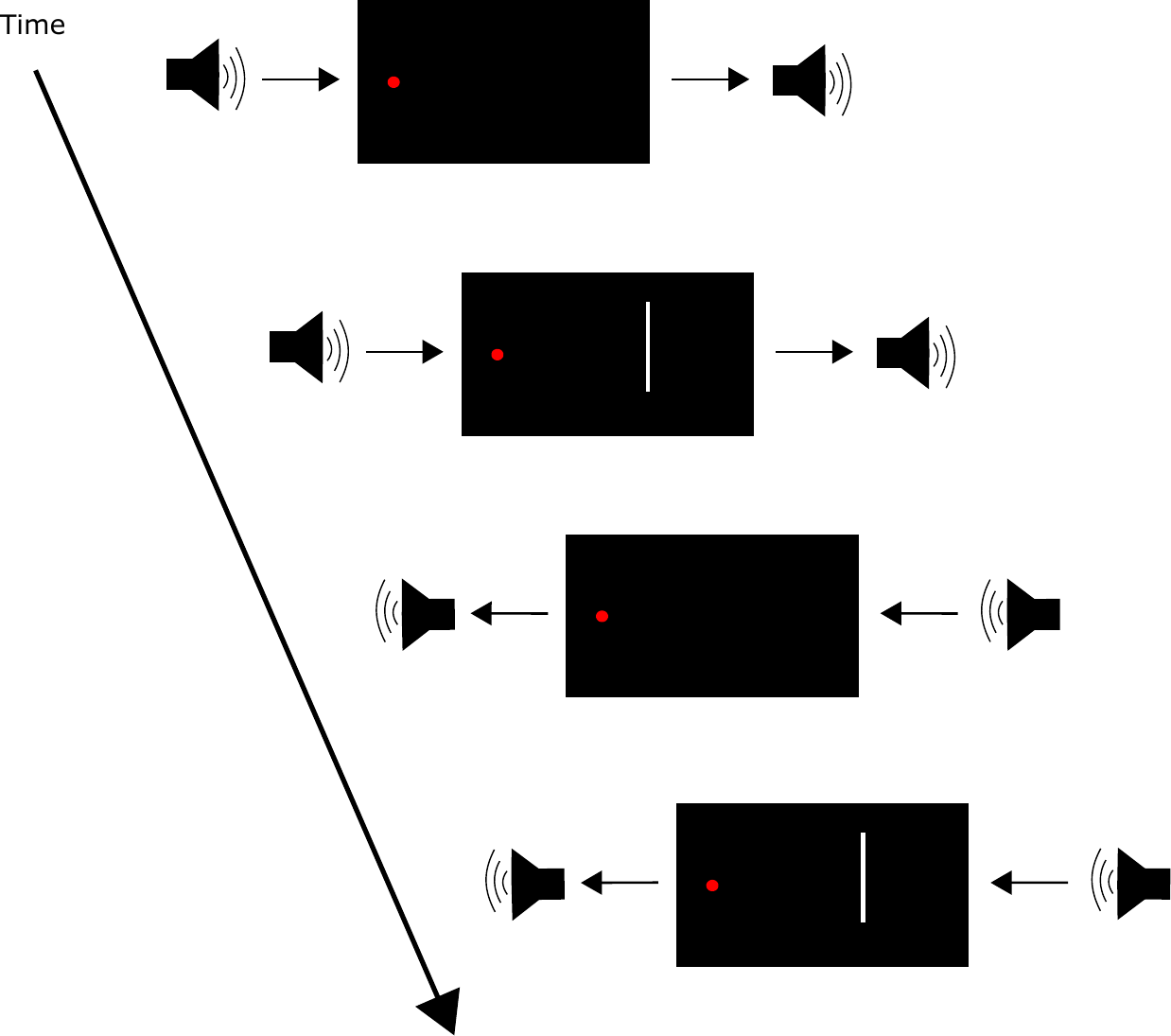}
	\end{center}
	\captionof{figure}{\label{fig:SIVM} Trial events in the sound induced illusory visual motion paradigm \protect\cite{Hidaka2011auditory}. The flashing visual stimulus (white bar above) was presented with varying eccentricities from fixation (red dot) depending on the trial condition. The auditory stimulus in the illusory condition was first presented to one ear and panned to the opposing ear. This paradigm resulted in the illusion of motion often in the same direction as that of the auditory stimulus motion. }
	
\end{figure}

Auditory motion can also have a profound effect on visual perception where a static flashing visual target is perceived to move in the same direction as auditory stimuli (Figure \ref{fig:SIVM}, see also \citeA{Hidaka2011auditory, Fracasso2013}). Perceived location of apparent motion visual stimuli can also be modulated by auditory stimuli \cite{Teramoto2012}. The visual motion direction selective brain region MT/V5 is activated in the presence of moving auditory stimuli, suggesting processing for auditory motion occurs there~\cite{Poirier2005}, which in turn hints at an intrinsic link between auditory and visual perceptual systems. These effects taken together again point to evidence accumulation in audio-visual integration as an optimal process, where different weight is given to auditory and visual inputs.

Consideration should be lent to how and when multiple stimuli in single modality should be grouped together as originating from a single source, or not, before pairing with another modality. As demonstrated above from neuropsychological evidences, and also from recent systems neuroscience evidences \cite{ghazanfar2006neocortex, meijer2019circuit}, the human audio-visual integration system appears to operate in rather complex processing steps, in addition to the traditional thinking of hierarchical processing from single modality. Hence, there is a need for modelling these cognitive processes. When designing artificial cognitive systems, efforts should be made to isolate sources of auditory and visual stimuli, and identify characteristics that would suggest they are related events. As shown above, relying on similar temporal signatures alone is not a robust approach when integrating signals across modalities.  The illusions discussed above are summarised in Table \ref{table:Illusion Table}.

\begin{landscape}
	\begin{table*}[h]
		
		\centering
		\renewcommand{\arraystretch}{1.5}
		
		\Huge
		
		\caption{\label{table:Illusion Table}
			Illusions summary. } 
		
		\resizebox{\textwidth}{!}{\input{Illusion_Table.tex}}
		
	\end{table*}
\end{landscape}

\section{Computational Cognitive Models}
We have discussed how auditory stimuli can have a pronounced effect on the perception of visual events, and vice versa, be it temporal or qualitative in nature. For auditory stimuli affecting the perception of visual signals, some effects were additive, facilitatory, and others suppressive. Regardless of the outcome, the influence of auditory stimuli on visual perception provides evidence of complexity of audio-visual integration processes on the way to visual perception. 

These complex mechanisms of how and when auditory stimuli alter visual perception have been clarified through computational modelling. \citeA{Chandrasekaran2017computational} presents a review of computational models of multisensory integration, categorizing the computational models into accumulator models, probabilistic models, or neural network models. These types of models are also typically used in single-modal perceptual decision-making (e.g. \citeA{Ratcliff1978theory, Wang2002probabilistic, Wong2006recurrent}). In this review, we will only focus on the accumulator and probabilistic models; the neural network (connectionist) models provide finer-grained, more biologically plausible description of neural processes, but on the behavioral level are mostly similar to the models reviewed here \cite{ma2006bayesian, Bogacz2006physics, Wong2006recurrent, roxin2008neurobiological, ma2008linking, liu2009dynamical, pouget2013probabilistic, ursino2014neurocomputational, zhang2016decentralized, ursino2019explaining, meijer2019circuit}.
\vspace{1em}

\subsection{Accumulator models}
The race model is a simple model that accounts for choice distribution and reaction time phenomena, e.g., faster reaction times of multisensory than unisensory stimuli \cite{Raab1962statistical, Gondan2016tutorial, Miller2016statistical}. More formally, the multisensory processing time $D_{AV}$ is the winner of two channel's processing times $D_{A}$ and $D_{V}$ for audio and visual signals: 
\begin{equation} 	
\label{eq:A}
D_{AV}=min(D_{A},D_{V}) \newline
\end{equation}

Another type of accumulator model, the coactivation model \cite{Schwarz1994diffusion, Diederich1995intersensory}, is based on the classic accumulator-type drift diffusion model (DDM) of decision-making~\cite{Stone1960models, Link1975relative, Ratcliff1978theory, Ratcliff1998modeling}. 	
The DDM is a continuous analogue of a random walk model \cite{Bogacz2006physics}, using a drift particle with state $X$ at any moment in time to represent a decision variable (relating in favour of one over another choice). This is obtained through integrating noisy sensory evidence over time in the form of a stochastic differential equation, a biased Brownian motion equation:
\begin{equation} 
\label{eq:B}
dX=Adt+cdW, \newline
\end{equation}
\noindent where $A$ is the stimulus signal (i.e. the drift rate), $c$ is the noise level, and $W$ represents the stochastic Wiener process. Integration of the sensory evidence begins from an initial point (usually origin point $0$), and is bounded by the lower and upper decision thresholds, $-z$ and $z$ respectively. Each threshold corresponds to a decision in favour of one of the two choices. Integration of the sensory information continues until the drift particle encounters either the upper or the lower threshold, at which stage a decision is made in favour of the corresponding option. The drift particle is then reset to the origin point to allow the next decision to be processed. The DDM response time (RT) is calculated as the time taken for the drift particle to move from its origin point to the either of the decision thresholds and can include a brief, fixed non-decision latency. For the simplest DDM, the RT has a closed form analytical solution \cite{Ratcliff1978theory, Bogacz2006physics}: 
\begin{equation} 
\label{eq:C}
RT=\frac{z}{A}\tanh(\frac{Az}{c^2}) \newline
\end{equation}
\noindent Similarly, the corresponding analytical solution for the DDM's error rate (ER) is:

\begin{equation} 
\label{eq:D}
ER=\frac{1}{1+\exp(\frac{2Az}{c^2})} \newline
\end{equation}

A simple unweighted coactivation model would combine evidence from two modalities and integrate it over time using the DDM \cite{Schwarz1994diffusion, Diederich1995intersensory}. For example, with unimodal sensory evidence $X_{1}$ and $X_{2}$ (e.g., auditory and visual information), the combined evidence is just a simple summation over time using (\ref{eq:B}): 

\begin{equation} 
\label{eq:E}
X_{c}=X_{1}+X_{2} \newline
\end{equation}

\subsection{Bayesian models}
In contrast to these models, the Bayesian modeling framework offers an elegant approach to modeling multisensory integration \cite{Angelaki2009multisensory}, although they share some similar characteristics with drift-diffusion models \cite{bitzer2014perceptual, fard2017bayesian}. This approach can provide optimal or near optimal integration of multimodal sensory cues by weighting the incoming evidences from each modality. For example, if the modalities follow a Gaussian distribution, the optimal integration estimate $X_{c}$ is \cite{Bulthoff1996bayesian}: 

\begin{equation} 
\label{eq:F}
X_{c}=\frac{k_{1}^2}{k_{1}^2+k_{2}^2}\widehat{X_{1}}+\frac{k_{2}^2}{k_{1}^2+k_{2}^2}\widehat{X_{2}} \newline
\end{equation}

\noindent where $k_{1}=\frac{1}{\sigma_{1}^2}$ and {$k_{2}=\frac{1}{\sigma_{2}^2}$} and $\widehat{X_{1}}$, $\widehat{X_{2}}$, $\sigma_{1}$ and $\sigma_{2}$ are the means and standard deviations for modality 1 and 2, respectively. Interestingly, the model can show that the combined variance (noise) will always be less than their individual estimates when the latter are statistically independent \cite{Alais2004ventriloquist}. This justifies why combining the signals help reduce the overall noise. In fact, \citeA{beck2012} makes a strong case for suboptimal inference, that the larger variability is due to deterministic, but suboptimal computation, and that the latter, not internal or external noise, is the major cause of variability in behaviour.

A more complex model, TWIN (time window of integration), involves a combination of the race model and the coactivation model~\cite{colonius2004multisensory}. Specifically, whichever modality is first registered (as in winning a ``race''), the size of the window is dynamically adapted to the level of reliability of the sensory modality. This would ensure, for instance, that if the less reliable modality wins the race, the window would be increased to give the more reliable modality a relatively higher contribution in multisensory integration. This model accounts for the illusory temporal order induced via a tone after visual stimuli onset, where the more reliable temporal information (auditory stimulus) dictates the perceptual outcome -- illusory temporal order \cite{Boyce2019}.  

The fission and fusion in audio-visual integration were suggested to result from statistically optimal computational strategy~\cite{Shams2010}, similar to Bayesian inference where audio-visual integration implies decisions about weightings assigned to signals and decisions whether to integrate these signals. \citeA{Battaglia2003bayesian} applied this Bayesian approach to reconcile two seemingly separate audio-visual integration theories. The first theory, called visual capture, is a ``winner-take-all'' model where the most reliable signal (least variance) dominates, while the second theory used a maximum-likelihood estimation to identify the weight average of the sensory input. The visual signal was shown to be dominant because of the subject perceptual bias, but the weighting given to auditory signals increased as visual reliability decreased. \citeA{Battaglia2003bayesian} showed that Equation~\ref{eq:F} can naturally account for both theories by having the weights to vary based on the signals' variances.

To study how children and adults differ in audio-visual integration, \citeA{Adams2016development} also used the same Bayesian approach in addition to two other models of audio-visual integration: a focal switching model, and a modality-switching model. The focal switching model stochastically sampled either auditory or visual cues based on subjects' reports of the observed stimulus. For the modality-switching model, the stochastically sampled cues were probabilistically biased towards the likelihood of the stimulus being observed. \citeA{Adams2016development} found that the sub-optimal switching models modeled sensory integration in the youngest study groups best. However, the older participants followed the partial integration of an optimal Bayesian model. 
\vspace{1em.}

\subsubsection{Illusions as a by-product of optimal Bayesian integration}
A variety of perceptual illusions have been shown to result from optimal Bayesian integration of information coming from multiple sensory modalities. In the context of sound-induced flash illusion (Figure~\ref{fig:DFI}), given independent auditory and visual sensory signals $A, V$, the ideal Bayesian observer estimates posterior probabilities of the number of source signals as a normalized product of single-modality likelihoods $P(A|Z_A)$ and $P(V|Z_V)$ and joint priors $P(Z_A, Z_V)$
	\begin{equation}
	\label{eq:opt_observer}
	P(Z_A, Z_V | A, V) = \frac{P(A|Z_A)P(V|Z_V)P(Z_A, Z_V)}{P(A,V)}.
	\end{equation}	
Regardless of the degree of consistency between auditory and visual stimuli, the optimal observer~(\ref{eq:opt_observer}) have been shown to be consistent with the performance of human observers~\cite{Shams2005optimal}. Specifically, when the discrepancy between the auditory and visual source signals is large, human observers rarely integrate the corresponding percepts. However, when the source signals overlap to a large degree, the two modalities are partially combined; in these cases the more reliable auditory modality shifts the visual percepts, thereby leading to sound-induced flash illusion.

Existence of different causes for signals of different modalities is the key assumption of the optimal observer model developed in~\cite{Shams2005optimal}, which allowed it to capture both full and partial integration of multisensory stimuli, with the latter resulting in illusions. \citeA{Kording2007causal} suggested that in addition to integration of sensory percepts, optimal Bayesian estimation is also used to infer the causal relationship between the signals; this was consistent with spatial ventriloquist illusion found in human participants. Alternative Bayesian accounts developed by~\citeA{Alais2004ventriloquist} (using Equation~\ref{eq:F}) and \citeA{Sato2007} also suggest that the spatial ventriloquist illusion stems from the near optimal integration of spatial and auditory signals.

Evidence for optimal Bayesian integration as the primary mechanism behind perceptual illusions comes from the paradigms involving not only audio-visual, but also other types of information. \citeA{wozny2008human} applied the model of \citeA{Shams2005optimal} to trimodal, audio-visuo-tactile perception, through simple extension of Equation~\ref{eq:opt_observer}. This Bayesian integration model accounted for cross-modal interactions observed in human participants, including touch-induced auditory fission, and flash- and sound-induced tactile fission~\cite{wozny2008human}. Further evidence for Bayesian integration of visual, tactile, and proprioceptive information is provided by the \textit{rubber hand illusion}~\cite{botvinick1998rubber}, in which a feeling of ownership of a dummy hand emerges soon after simultaneous tactile stimulation of both the concealed own hand of the participant and the visible dummy hand (see \citeA{lush2020demand} for a critique of control methods used in the `rubber hand' illusion). The optimal causal inference model~\cite{Kording2007causal} adapted for this scenario accounted for this illusion~\cite{samad2015perception}. Moreover, the model predicted that if the distance between the real hand and the rubber hand is small, the illusion would not require any tactile stimulation, which was also confirmed experimentally~\cite{samad2015perception}.

Finally, Bayesian integration has recently been shown to account even for those illusions which were previously striking researchers as ``anti-Bayesian'', for the reason that the empirically observed effects had the direction opposite to the effects predicted by optimal integration. Such ``anti-Bayesian'' effects are the size-weight illusion~\cite{peters2016sizeweight} (of the two objects with same mass but different size, the larger object is perceived to be lighter), and the material-weight illusion~\cite{peters2018materialweight} (of the two objects with the same mass and size, the denser-looking object is perceived to be lighter). In both cases, the models explaining these two illusions involved optimal Bayesian estimation of latent variables (e.g., density), which affected the final estimation of weight. 

Altogether, the reviewed evidence from diverse perceptual tasks illustrates the ubiquity of optimal Bayesian integration and its role in emergence of perceptual illusions.
\vspace{1em}

\subsubsection{Temporal dimension in Bayesian integration}
Basic Bayesian modelling framework often does not come with a temporal component, unlike dynamical models such as accumulator. However, a recent study shows that when optimal Bayesian model is combined with the DDM, it can provide optimal and dynamic weightings to the individual sensory modalities. In the case of visual and vestibular integration, using an experimental setup similar to that of \citeA{Fetsch2009dynamic}, \citeA{Drugowitsch2014optimal} found a Bayes-optimal DDM to integrate vestibular and visual stimuli in a heading discrimination task. It allowed the incorporation of time-variant features of the vestibular motion, i.e. motion acceleration, and visual motion velocity. The Bayesian framework allowed the calculation of a combined sensitivity profile $d(t)$ from the individual stimulus sensitivities.

\begin{equation} 
\label{eq:G}
d(t)=\sqrt{\frac{k_{vis}^2(c)}{k_{comb}^2(c)}{v^2}(t)+\frac{k_{vest}^2(c)}{k_{comb}^2(c)}{a^2}(t)} \newline
\end{equation}
\noindent where $k_{vis}(c)$, $k_{vest}(c)$, and $k_{comb}(c)$ are the visual, vestibular and combined stimulus sensitivities, and $v(t)$ and $a(t)$ are the temporal sensitivities of the visual and vestibular stimuli, respectively. \citeA{Drugowitsch2014optimal} found that Bayes-optimal DDM led to suboptimal integration of stimuli when subject response times were ignored. However, when response times were considered, the decision-making process took longer but resulted in more accurate responses. That said, a significant limitation of the study by \citeA{Drugowitsch2014optimal} and related work is that it does not incorporate delays in information processing. More generally, current Bayesian models do not consider how temporal delays impact sensory reliability. Delays are particularly relevant for feedback control in the motor system and processes like audio-visual speech because different sensory systems are affected by different temporal delays \cite{mcgrath1985intermodal, jain2015comparative, Crevecoeur2016}.

So far, the modeling approaches do not generally take into account the effects of attention, motivation, emotion, and other ``top-down'' or cognitive control factors that could potentially affect multimodal integration. However, there are experimental studies of top-down influences, mainly attention \cite{Talsma2010multifaceted}. More recently, \citeA{Maiworm2012emotional} showed that aversive stimuli could reduce the ventriloquism effect. \citeA{Bruns2014reward} designed a task paradigm in which rewards were differentially allocated to different spatial locations (hemifields), creating a conflict between reward maximization and perceptual reliance. The auditory stimuli were accompanied by task-irrelevant, spatially misaligned visual stimuli. They showed that the hemifield with higher reward had a smaller ventriloquism effect. Hence, reward expectation could modulate multimodal integration and illusion, possibly through some cognitive control mechanisms. Future computational studies, e.g., using reward rate analysis \cite{Bogacz2006physics, niyogi2013dynamic}, should address how reward and punishment are associated with such effects. 

\section{Audio-visual systems in the artificial}
Multimodal integration and sensor fusion in artificial systems have been an active research field for decades \cite{luo89multisensor, luo02multisensor}, since using multiple sources of information can improve artificial systems in many application areas, including smart environments, automation systems and robotics, intelligent transportation systems. Integration of sensory modalities to generate a percept can occur at different stages, from low (feature) to high (semantic) level. The integration of several sources of unimodal information at middle and high level representations \cite{gomez16multimodal, wu99multimodal} has clear advantages: interpretability, simplicity of system design, and avoiding the problem of increasing dimensionality of the resulting integrated feature. Although model dependent, lower dimensionality of the feature space typically leads to better estimates of parametric models and computationally faster non-parametric models for a fixed amount of training data, which in turn can reduce the number of judgement failures. However, percept integration at the representation level lacks robustness and does not account for the way humans integrate multisensory information \cite{calvert2001detection, shams2005early, stein2014development, watkins2006} to create these percepts \cite{cohen01multimodal}. Temporal ventriloquism and the McGurk effect are just two examples of the result of the lower-level integration of sensory modalities in humans to create percepts, yet the differences with artificial systems go even	further. While human perceptual decision-making is based on a dynamic process of evidence accumulation of noisy sensory information over time (see above), artificial systems typically follow a snapshot approach, i.e. percepts are created on the basis of instantaneous information, and only from data over time-windows when the perception mainly unfolds over time. Therefore, we can distinguish between decisions made over accumulated evidence, i.e. decision-making, and decisions made following the snapshot approach, i.e. classification, even though sometimes these two approaches are combined.

Audio-visual information integration is one of the multi-sensory mechanisms that has increasingly attracted research interests in the design of artificial intelligent systems. This is mainly due to the fact that humans heavily rely on these sensing modalities, and advances in this area have been facilitated by the high level of maturity of the individual areas involved, for instance signal processing, speech recognition, machine learning, and computer vision. See \citeA{parisi2017computational} and \citeA{parisi2018neurorobotic} for examples of how human multisensory integration in spatial ventriloquism has been used to model human-like spatial localisation responses in artificial systems in which -- given a scenario where sensor uncertainty exists in audio-visual information streams -- they propose artificial neural architectures for multisensory integration. An interesting characteristic of audio-visual processing compared to other multimodal systems is the fact that the information unfolds over time for audio signals, but also for visual systems when video is considered instead of still images. However, most of the research in artificial visual systems follow the snapshot approach mentioned above to build percepts, while video processing mainly focuses on integrating and updating of these instantaneous percepts over time, which can be seen as evidence accumulation. Like for other multimodal integration modalities, audio-visual integration in artificial systems can be performed at different levels, although is generally used for classification purposes, while decision-making, when performed, is based on the accumulation of classification results. Optimal temporal integration of visual evidence together with audio information can be prone to the sort of illusory effects on percepts illustrated above in humans. However, because artificial systems are designed with very specific objectives, an emergent deviation of the measurable targets of the system would be considered as a failure or bug of the system. Therefore, although artificial systems can display features that could be the emergent results of the multimodal integration, they will be regarded as failures to be avoided, and most likely not reported in the literature. A close example related to reinforcement learning is the reward hacking effect \cite{Amodei2016concrete}, where a learning agent finds an unexpected (maybe undesired) optimal policy for a given learning problem.

As stated earlier, multimodal integration is typically performed at high level, as low-level integration generates higher dimensional data, thereby increasing the difficulty of processing and analysis. Moreover, the low-level integration of raw data can have the additional problem of combining data of very different nature. The dimensionality problem is magnified by the massive amount of data visual perception produces, therefore most approaches to audio-visual processing in intelligent systems also address the problem of integration at a middle and higher levels across diverse applications: object and person tracking \cite{beal03graphical, nakadai02real}, speaker localization and identification \cite{gatica07audiovisual}, multimodal biometrics \cite{chibelushi02review}, lip reading and speech recognition \cite{perez05lip, sumby54visual, luettin97speechreading, chen10audiovisual} and video annotation \cite{li04content, wang00multimedia}, and others. The computational models described above can be identified with these techniques for artificial systems, as they generate percepts and perform decision-making on the basis of middle-level fusion of evidence.  However, some work in artificial systems deals with the challenging problem of combining data at the signal level \cite{fisher00learning, fisher04speaker}. While artificial visual systems were dominated by feature definition, extraction, and learning \cite{li08comprehensive}, the success of deep learning and convolutional neural networks in particular has shifted the focus of computer vision research. Likewise, speech processing is adopting this new learning paradigm, yet audio-visual speech processing with deep learning is still based mainly on high-level integration \cite{noda15audio, deng13new}. Although the human audio-visual processing is not fully understood, our knowledge of the brain strongly inspires (and biases) the design of artificial systems. Besides the well-known (yet not widely reported) reward hacking in reinforcement learning and optimisation \cite{Amodei2016concrete}, to the best of our knowledge no illusory percepts have been reported in specific-purpose artificial systems, as they are typically situations to be avoided.

\section{Discussion}
The multi-modal integration processes and related illusions outlined above are closely related in terms of how audition affects visual perception. Untangling whether prior entry (whatever form it may take), impletion, temporal ventriloquism, or featural similarity of auditory stimuli are the drivers of audio-visual effects can be a challenge, and may be missing the bigger picture when trying to understand how perception is arrived at in a noisy world. The most likely explanation of the discussed effects is one of an overarching unified process of evidence accumulation and evidence discounting. This perspective would state that evidence is gathered via multiple modalities and is filtered through multiple sub-processes: prior entry, auditory streaming, impletion, and temporal ventriloquism. Two or more of these sub-processes will often interact, with various weightings given to each process. For example, prior entry using a single auditory cue can induce an illusion of temporal order, but with the addition of a cue in the unattended side of space after the presentation of both target visual stimuli, extra information in favour of temporal order can be accumulated, which would increase the strength of the illusion \cite{Boyce2019}. Similarly, illusory temporal order can be induced via spatially neutral tones (an orthogonal design), as demonstrated by \citeA{Boyce2019}, which appears to combine prior entry and temporal ventriloquism, and impletion-like processes generally. The illusory temporal order induced via spatially neutral tones is significantly weaker when compared to the spatially congruent audio and visual stimuli equivalent, highlighting the relative weight given to spatial congruency. Additionally, when featurally distinct tones are used for both of these effects, the prior entry illusory order is preserved while the illusory order induced by spatially neutral tones is completely abolished \cite{Boyce2019}. This highlights how spatial information carries greater weight than the featural information of the tones used when the auditory and visual stimuli are spatially congruent. Conversely, it also highlights how featural characteristics carry greater weight in the absence of audio-visual spatial congruency.

As outlined above, temporal ventriloquism effects can also interact with auditory streaming where the features of auditory stimuli undergo a process of grouping, and the outcome can dictate whether the stimuli is paired or not with visual stimuli. The mere fact that the temporal signature of stimuli is not in-and-of-itself enough to induce an effect (at the times discussed here) suggests that sub-processes interact across modalities. Specifically, when auditory stimuli are not grouped in the streaming process, there is less evidence that they belong to the same source, and in turn it is less likely both auditory stimuli belong to the same source as the visual events. These types of interactions taken with different outcomes in visual perception, depending on the number of auditory stimuli used, point towards an overarching process that fits an expanded version of impletion, or a unifying account of impletion \cite{Boyce2019} (aligning with Bayesian inference), where the most likely real world outcome is reflected in perception. The observer weighs evidence from both modalities in multi-modal perception and also weighs evidence within a single modality. This suggests an inherent weighted hierarchy, where spatial, featural, and temporal information are all taken into account.

The discussed integration processes are often statistically optimal in nature~\cite{Shams2010, Alais2004ventriloquist}. This has implications for designing artificial cognitive systems. An optimal approach may be an intuitive one: minimising the average error in perceptual representation of stimuli. However, as discussed, this approach can come with costs in terms of illusions, or artefacts, despite a reduction in the average error. Of course, some systems will not rely wholly on mimicking human integration of modalities, and indeed will supersede human abilities: for example, a system may be designed to perform multiple tasks simultaneously, something a human cannot do. However, future research should aim to identify when an optimal approach is not suitable in multi-modal integration.

Using illusion research in human perception as a guide, researchers could identify and model when artefacts occur in multi-modal integration, and apply these findings to system design. This might take the shape of modulating the optimality of integration depending on conditions via increasing or decreasing weightings as deemed appropriate. This approach could contribute to a database of ``prior knowledge'' where specific conditions that can result in artefacts are catalogued and can inform the degree of integration between sources in order to avoid undesired outcomes. For instance, \citeA{Roach2006resolving} examines audio-visual integration from just such a perspective using a Bayesian model of integration, where prior knowledge of events are taken into account and a balance between benefits and costs (optimal integration and potential erroneous perception) of integration is reached. They examined interactions between auditory and visual rate perception (where a judgement is made in a single modality and the other modality is `ignored') and found that there is a gradual transition between partial cue integration and complete cue segregation as inter-modal discrepancy increases. The Bayesian model they implemented took into account prior knowledge of the correspondence between audio and visual rate signals, when arriving at an appropriate degree of integration.

Similarly, a comparison between unimodal information and the final multi-modal integration might offer a strategy for identifying artefacts. This strategy might be akin to the study of \citeA{Sekiyama1994}, which demonstrated that Japanese participants, in contrast to their American counterparts, have a different audio-visual strategy in the McGurk paradigm: less weight was given to discrepant visual information, which in turn affected the integration with auditory stimuli, ultimately resulting in a smaller McGurk effect. The inverse was shown in the participants with cochlear implants who demonstrated a larger McGurk effect: more weight was given to visual stimuli in general \cite{Rouger2008}. \citeA{Magnotti2017} suggested that a causal inference (determining if audio and visual stimuli have the same source) ``type'' calculation is a step in multisensory speech perception, where some, but not all, incongruent audio-visual speech stimuli are integrated based on the likelihood of a shared, or separate, sources. Should that be the case, and this step is part of a near optimal strategy, a suboptimal process --- such as a comparison of unimodal information and final multi-modal integration, or adjusting relative stimulus feature weightings when estimating likelihoods of source --- could ensure that a McGurk-like effect is avoided. Additionally, it is worth noting the research by \citeA{driver1996enhancement} who demonstrated that when there are competing auditory speech stimuli ostensibly from the same source and a matching visual speech stimuli from a different spatial location this has the effect of `pulling' the matching auditory stimuli in perceptual space towards the visual stimuli improving separation of the auditory streams reflected in report accuracy.

Dynamic adjustment of prior expectations is a vital consideration when designing local ``prior knowledge'' databases for artificial cognitive systems. This is illuminated by the fact that dynamically updated prior expectations can increase the likelihood of audio-visual integration: When congruent audio-visual stimuli is interspersed with incongruent McGurk audio-visual stimuli, the illusory McGurk effect emerges~\cite{Gau2016}. Essentially, when there is a high instance of audio-visual integration due to congruent stimuli, incongruent stimuli have a greater chance of being deemed as originating from the same source and therefore being integrated. These behavioural results were supported by fMRI recordings that showed the left inferior frontal sulcus arbitrates between multisensory integration and segregation by combining top-down prior congruent/incongruent expectations with bottom-up congruent/incongruent cues \cite{Gau2016}. This suggests that in artificial cognitive systems, even though the prior knowledge databeses should cater for updates, it should not be done so live and ``in the wild''. If the probability of audio and visual stimuli originating from the same source was calculated near-optimally in the manner described by \citeA{Gau2016}, dynamically it could result in artefacts in an artificial cognitive system, where, for example, unrelated audio-visual events could be classified as being characteristics of the same event. If a dynamic approach is required, an optimal strategy should be avoided for these reasons.

In addition to the approaches suggested above, it is important that temporal characteristics, such as processing differences across artificial modalities are also taken into account. For instance, even though light is many hundreds of thousands of times faster than sound, the human perceptual system processes sound stimuli faster than visual stimuli \cite{recanzone2009interactions}. Indeed, it has been suggested that characteristics such as processing speeds of auditory and visual stimuli changing as a person ages (for example, visual processing slowing) may be responsible for increasing audio-visual integration in older participants where auditory tones had a greater influence on the perceived number of flashes in the sound-induced flash illusion compared to younger participants \cite{deloss2013multisensory, mcgovern2014sound}. Similar considerations should be made for artificial systems. Regardless of how sensitive or fast at processing a given artificial sensor is, light will always reach a sensor before sound if the respective stimuli originate from the same distance/location. Setting aside the physical attribute of the speed of light versus the speed of sound, there is an additional level of complexity even in an artificial system where it presumably would require a lot more computational power, and thus time, to process and separate the stimulus of interest in a given visual scene (with other factors such as feature resolution playing a role). Indeed, as mentioned previously, temporal feedback delay in the nervous system is a factor in optimal multi-sensory integration \cite{Crevecoeur2016}. Additionally, a unimodal auditory strategy for separating sources of auditory stimuli in a noisy environment via extracting and segregating temporally coherent features into separate streams has been developed by \cite{Krishnan2014}. These considerations taken with the multi-modal audio-visual strategies deployed in speech (where temporal relationships of mouth movement and auditory onset play a role, specifically the voice onsets between 100 and 300ms before the mouth visibly moves \cite{Chandrasekaran2009}) highlight the importance of temporal characteristics, correlations, and strategies when designing artificial cognitive systems. Finally, to handle noisy sensory information, artificial cognitive systems should perhaps consider incorporating temporal integration of sensory evidence \cite{yang2017single, rano2017drift, mi2019spatiotemporal} instead of employing snapshot decision processing.

In summary, we highlight a wide range of audio-visual illusory percepts from the psychological and neuroscience literature, and discussed how computational cognitive models can account for some of these illusions -- through seemingly optimal multimodal integration. We provide cautions regarding the na\"{\i}ve adoption of these human multimodal integration computations for artificial cognitive systems, which may lead to unwanted artefacts. Further investigations of the mechanisms of multimodal integration in humans and machines can lead to efficient approaches for mitigating and avoiding unwanted artefacts in artificial cognitive systems.

\section*{Conflict of Interest Statement}
The authors declare that the research was conducted in the absence of any commercial or financial relationships that could be construed as a potential conflict of interest.

\section*{Author Contributions}
WPB wrote the Introduction, Audio-visual Integration, and Discussion sections. AL, AZ, and KFW-L wrote the Computational Cognitive Models section. IR wrote the Audio-visual Systems in the Artificial section. All authors contributed to structuring, revising, and proofreading of the manuscript.

\section*{Funding}
This research was supported by the Northern Ireland Functional Brain Mapping Project (1303/101154803), funded by Invest NI and the University of Ulster (KFW-L), The Royal Society (KFW-L, IR), and ASUR (1014-C4-Ph1-071) (KFW-L, IR, AL). 

\section*{Acknowledgements}
The authors are also grateful to Shufan Yang for helpful discussions on the computational modelling.

\bibliography{references}
\end{document}

%% file: Illusion_Table.tex
\begin{tabulary}{\textwidth}{ l@{\hskip 0.5in} l@{\hskip 0.5in} l@{\hskip 0.5in} l@{\hskip 0.5in} l@{\hskip 0.5in} l@{\hskip 0.5in} l@{\hskip 0.5in} l@{\hskip 0.5in} l@{\hskip 0.5in} l@{\hskip 0.5in}}
\\
\hline
\\
Illusion  & Description \\
\hline
\\


The line-motion illusion: & When one side of space is cued prior to the presentation of the entire physical line it results in\\ &the perception of the line being drawn from that cued side of space  \\ \\

Illusory temporal order I: & When a tone is presented to one side of congruent space prior to the simultaneous\\ &presentation of both targets in a simultaneity judgement task, illusory sequential order is perceived \\ \\

Illusory temporal order II: & When a tone is presented in neutral space prior to the simultaneous presentation of both targets in\\ &a ternary judgement task, and a tone is presented in neutral space after the onset of both targets, illusory sequential\\ &order is perceived \\ \\

Temporal ventriloquism -
performance enhancement: & When a tone is presented before the first visual stimulus in a temporal order judgement sequence and a tone\\ &is presented after the second visual stimulus, performance is improved \\ \\

Temporal ventriloquism -
performance detriment I: & When a tone is presented after the first visual stimulus in a temporal order judgement sequence and a tone \\ &is presented before the second visual stimulus, performance is worsened \\ \\

Temporal ventriloquism -
performance detriment II: & When a single tone is presented before the first visual stimulus in a temporal order judgement sequence,\\&response bias matching the presentation order of visual stimuli is reduced \\ \\

Temporal ventriloquism -
illusory apparent visual motion: & When auditory stimuli are presented in neutral space but with specific SOAs in relation to visual\\ &stimuli the perception of apparent motion can be modulated \\ \\

Multiple flash illusion: & When two tones are presented either side of a single presentation of a circle in time, multiple flashes \\ &of the circle are perceived \\ \\

Single flash illusion: & When a single tone is presented with multiple flashes of a circle a single presentation of the circle is \\ &perceived \\ \\

Sound-induced illusory apparent visual motion: & When auditory stimuli are presented panning from one ear to the other in time with a flashing\\ &visual target, visual apparent motion is perceived \\ \\

\end{tabulary}